\documentclass[letterpaper]{article} 
\usepackage{aaai2026}  
\usepackage{times}  
\usepackage{helvet}  
\usepackage{courier}  
\usepackage[hyphens]{url}  
\usepackage{graphicx} 
\usepackage{natbib}  
\usepackage{caption} 
\usepackage{algorithm}
\usepackage{algorithmic}

\usepackage{amssymb}
\usepackage{amsmath}
\usepackage{booktabs}
\usepackage{multirow} 
\usepackage{newfloat}
\usepackage{listings}
\DeclareCaptionStyle{ruled}{labelfont=normalfont,labelsep=colon,strut=off} 
\floatstyle{ruled}
\newfloat{listing}{tb}{lst}{}
\floatname{listing}{Listing}
\title{Enhancing Interpretability for Vision Models via Shapley Value Optimization}
\author{
    Kanglong Fan\textsuperscript{\rm 1},
    Yunqiao Yang\textsuperscript{\rm 1},
    Chen Ma\textsuperscript{\rm 1}\thanks{Corresponding author.}
}
\affiliations{
    \textsuperscript{\rm 1}Department of Computer Science, City University of Hong Kong\\

    kanglofan2-c@my.cityu.edu.hk, yqyang.cs@my.cityu.edu.hk, chenma@cityu.edu.hk
}

\usepackage{bibentry}

\begin{document}

\maketitle

\begin{abstract}
Deep neural networks have demonstrated remarkable performance across various domains, yet their decision-making processes remain opaque. Although many explanation methods are dedicated to bringing the obscurity of DNNs to light, they exhibit significant limitations: post-hoc explanation methods often struggle to faithfully reflect model behaviors, while self-explaining neural networks sacrifice performance and compatibility due to their specialized architectural designs. To address these challenges, we propose a novel self-explaining framework that integrates Shapley value estimation as an auxiliary task during training, which achieves two key advancements: 1) a fair allocation of the model prediction scores to image patches, ensuring explanations inherently align with the model's decision logic, and 2) enhanced interpretability with minor structural modifications, preserving model performance and compatibility. Extensive experiments on multiple benchmarks demonstrate that our method achieves state-of-the-art interpretability. 
\end{abstract}


\section{Introduction}

Deep neural networks (DNNs) have achieved remarkable success across numerous applications~\cite{redmon2016you,vinyals2015show,antol2015vqa}. Despite their impressive capabilities, a significant challenge persists: the inherent lack of interpretability in their decision-making processes. This limitation raises critical concerns about the reliability and safety of DNNs, particularly in high-stakes applications where model interpretability is crucial for ensuring reliability and accountability~\cite{molnar2020interpretable, borys2023explainable}.

Current approaches for explaining DNN predictions can be broadly categorized into two paradigms: post-hoc explaining and self-explaining methods. Post-hoc explanation methods, including gradient-based~\cite{sundararajan2017axiomatic,Selvaraju2017gradcam,yang2023idgi,li2023negative}, perturbation-based techniques~\cite{petsiuk2018rise,fong2019understanding,jethani2021fastshap, covert2023learning}, counterfactual-generation-based~\cite{bass2022icam,xie2024accurate} and attention-based~\cite{abnar2020quantifying, chefer2021generic, qiang2022attcat, wu2024token}, are typically applied independently of model training. While widely adopted, these methods often produce unfaithful explanations that inadequately represent model behaviors~\cite{adebayo2018sanity,yang2019benchmarking,kindermans2019reliability,hesse2024benchmarking}. In contrast, self-explaining neural networks (SENNs)~\cite{chen2019looks,brendel2019approximating, wang2021shapley,hesse2021fast,bohle2022bcos,chen2023harsanyinet,nauta2023pip,de2024patch,arya2025bcosification} integrate interpretability directly into their models through specialized architecture designs. By construction, SENNs generate intrinsic explanations aligned with the model’s decision logic, offering greater faithfulness compared to post-hoc methods. However, SENNs face three major limitations: 1) they often require training from scratch, limiting their compatibility with pre-trained models~\cite{arya2025bcosification}; 2) their specialized architectural designs~\cite{chen2019looks, hesse2021fast,chen2023harsanyinet} often lead to degraded performance compared to standard DNNs; 3) the inclusion of interpretability modules~\cite{brendel2019approximating, wang2021shapley, chen2023harsanyinet} introduces memory and computational overhead, hindering scalability.

\begin{figure*}
\centering
\includegraphics[width=1\textwidth]{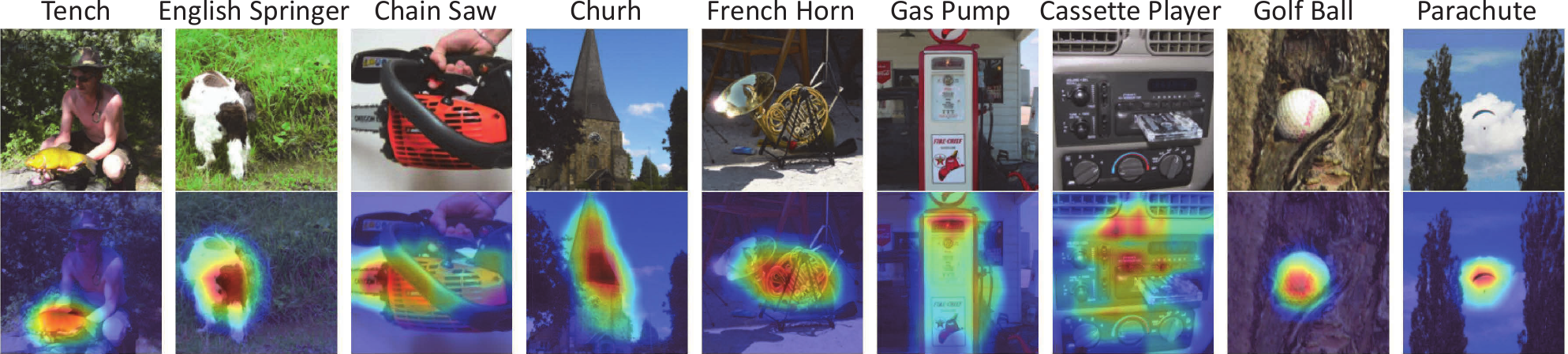}
\caption{Visualization of our method's explanations on ImageNet. The top row shows the input images and the bottom displays the corresponding explanations.}
\label{fig:1}
\end{figure*}

Recent research~\cite{lundberg2017unified,jethani2021fastshap,covert2023learning} establishes the Shapley value~\cite{shapley1953value} as a principled approach for model interpretation, as it quantifies the marginal contribution of individual input components (e.g., image patches) to predictions. While many approaches have made progress in incorporating Shapley value for model interpretation, they still face significant limitations in computational efficiency and attribution accuracy. Conventional approaches~\cite{castro2009polynomial, strumbelj2010efficient, lundberg2017unified, covert2021improving, mitchell2022sampling} require extensive model inferences to approximate Shapley values, making them computationally costly and impractical for many applications. Moreover, the discrepancy between masked data used for Shapley value estimation during testing and the unmasked data during training may introduce additional attribution errors. Alternative methods such as FastSHAP~\cite{jethani2021fastshap} and ViTShapley~\cite{covert2023learning} introduce an auxiliary surrogate model to process masked images, followed by training an explainer to fastly estimate Shapley values. However, the explainer ultimately interprets the surrogate model rather than the model to be explained.

\begin{figure*}
\centering
\includegraphics[width=0.95\textwidth]{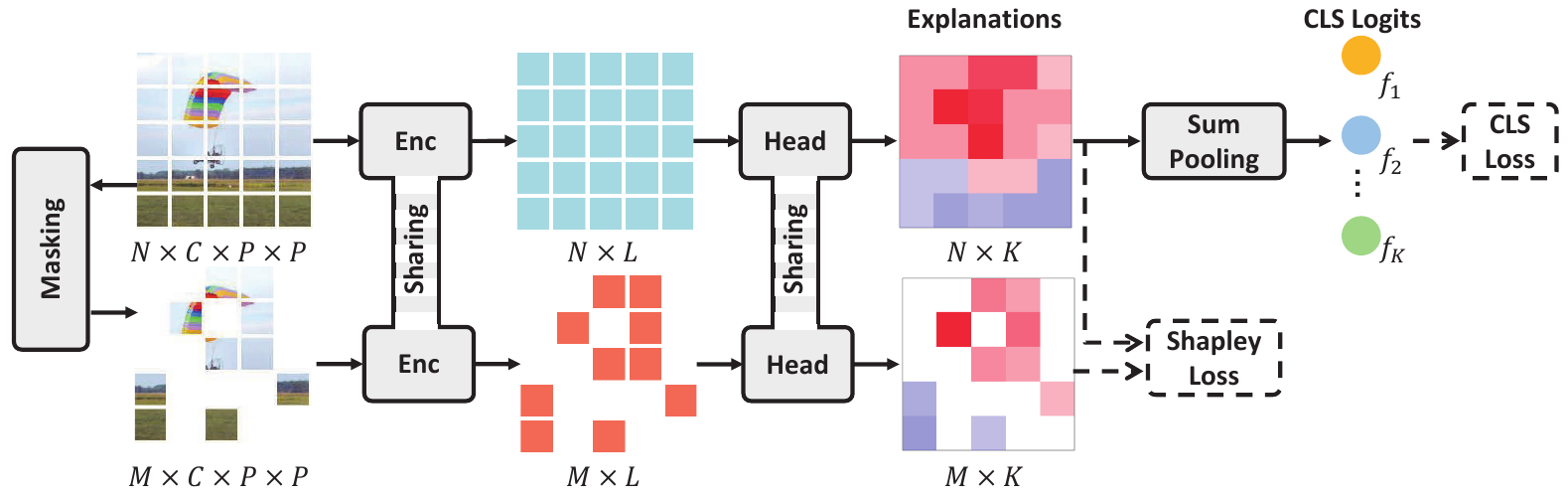}
\caption{Schematic representation of our proposed dual-branch self-explaining framework. The upper branch processes the original image to get the explanations and classification logits, while the lower branch processes the masked image to supervise the explanations via Shapley Loss.}
\label{fig:4}
\end{figure*}

To address these challenges, we propose a multitask learning framework that intrinsically integrates Shapley value estimation directly into the model’s optimization process. By simultaneously optimizing the model for the primary task (e.g., image classification) and Shapley value estimation with an appropriate trade-off parameter, our framework (see Figure~\ref{fig:4}) achieves self-interpretable predictions without compromising the performance of the primary task. Unlike post-hoc explanation methods that risk misrepresenting a model’s decision logic, our approach learns explanations as part of the model’s reasoning process, ensuring alignment between explanations and decision logic (see Figure~\ref{fig:1}). Compared to existing SENNs, our framework requires minor architectural modifications and does not rely on external interpretation modules, preserving computational efficiency and model compatibility. Furthermore, by naturally incorporating masked images during training, our approach circumvents the discrepancy problem associated with post-hoc Shapley estimation without the need for a surrogate model.

In summary, our key contributions are as follows:
\begin{itemize}
    \item we propose a novel optimization framework that jointly learns Shapley value estimation and the primary task, enhancing explanation faithfulness while preserving the primary task performance;
    \item we validate our method on image classification and image-text matching tasks, and demonstrate that it outperforms state-of-the-art post-hoc explaining and self-explaining methods.
\end{itemize}

\section{Related Work}

\label{sec:review}
Contemporary explainable artificial intelligence encompasses post-hoc explaining and self-explaining methods~\cite{nauta2023explainable}. Post-hoc methods including gradient-based approaches~\cite{sundararajan2017axiomatic,Selvaraju2017gradcam,yang2023idgi,li2023negative}, perturbation-based techniques~\cite{ribeiro2016should, lundberg2017unified, petsiuk2018rise,jethani2021fastshap,covert2023learning}, counterfactual-generation-based~\cite{bass2022icam, xie2024accurate} and attention-based~\cite{abnar2020quantifying, chefer2021generic,qiang2022attcat,wu2024token}, often face challenges with reliability and faithfulness~\cite{adebayo2018sanity,yang2019benchmarking,hesse2024benchmarking}. Among perturbation-based methods, conventional Shapley value approaches~\cite{castro2009polynomial, strumbelj2010efficient, lundberg2017unified, covert2021improving, mitchell2022sampling} are constrained by high computational complexity, while existing acceleration techniques~\cite{jethani2021fastshap, covert2023learning} often fail to produce accurate Shapley value estimates.

Models with inherent interpretability~\cite{chen2019looks,brendel2019approximating,hesse2021fast,nauta2021neural,bohle2022bcos,nauta2023pip,de2024patch} offer a more promising direction by incorporating transparency mechanisms into their architecture. ProtoPNet~\cite{chen2019looks}, PIPNet~\cite{nauta2023pip} and PIMPNet~\cite{de2024patch} leverages prototype comparison, BagNets~\cite{brendel2019approximating} employs patch-based feature extraction, $\mathcal{X}$-DNNs~\cite{hesse2021fast} enables efficient Integrated Gradients computation, and B-cos networks~\cite{bohle2022bcos} implement structural alignment for feature identification. Recent advances~\cite{wang2021shapley,chen2023harsanyinet} seek to incorporate Shapley value into self-explaining architectures. While enhancing the interpretability, the specialized architecture designs could increase memory and computational overhead, and compromise the model's compatibility and performance. B-cosified~\cite{arya2025bcosification} extends compatibility with pre-trained backbone models such as ResNet~\cite{he2016deep}, and Vision Transformers (ViT)~\cite{dosovitskiy2020image}. Although with minor architectural changes, the recent work~\cite{alkhatib2025prediction} primarily focuses on tabular data applications and small-scale classification tasks. Our method integrates Shapley value attribution directly into the learning process as an auxiliary task, tailored for fine-tuning modern, large-scale pretrained models like ViT and CLIP. We demonstrate its effectiveness and superior explanation quality across multiple tasks, including image classification, object localization (segmentation), and image-text matching on large-scale benchmarks.

\section{Background}


\subsection{Shapley Value}

Consider a feature space denoted by $\mathcal{X}$, where each observation $\mathbf{x}$ comprises $N$ distinct components expressed as $\mathbf{x} = (\mathbf{m}_1, \ldots, \mathbf{m}_N)$. In a classification framework, the target class $y$ is drawn from the set $\mathcal{Y} = \{1, \ldots, K\}$. The predictive model is represented by $\mathbf{f}(\mathbf{x}; \theta)$, where $\theta$ represents the model parameters and $f_y(\mathbf{x}; \theta)$ corresponds to the logit value for the class $y$.  A binary mask $\mathbf{s} \in \{0, 1\}^N$ indicates the selection of components from $\mathbf{x}$, where the subset $\mathbf{x}_\mathbf{s}=\{\mathbf{m}_i | s_i =1, i = 1,\dots, N\}$. The vectors $\mathbf{1}$ and $\mathbf{0}$ in $\mathbb{R}^N$ represent all-ones and all-zeros vectors, respectively, while $\mathbf{e}_i \in \mathbb{R}^N$ represents the $i-$th canonical basis vector (a one-hot vector with 1 at the $i-$th position and 0 elsewhere). $\mathbf{1}^\top \mathbf{s}$ yields the number of components in a subset.  Given an image $\mathbf{x}$,  the value function $v: 2^N \mapsto \mathbb{R}$ quantifies the predictive value of the subset $\mathbf{x}_\mathbf{s}$ for class $y$:
\begin{equation} \label{eq:value function}
    v_y(\mathbf{s}) = f_y \left(\mathbf{x}_\mathbf{s}; \theta \right).
\end{equation}

Within cooperative game theory, the Shapley value provides a principled method to fairly allocate collective gains among coalition participants \cite{shapley1953value}. Given the value function $v$ that maps subsets of participants to their aggregated contributions, the Shapley value $\phi_i$ quantifies the marginal contributions of component $i$, computed as:
\begin{equation}
\phi_i =  \sum_{\mathbf{s} | s_i = 0} \frac{(\mathbf{1}^\top \mathbf{s})!  (N-1-\mathbf{1}^\top \mathbf{s})!}{N!} \big( v(\mathbf{s} + \mathbf{e}_i) - v(\mathbf{s}) \big). \label{eq:shapley}
\end{equation}
Here, the summation extends over all masks $\mathbf{s}$ excluding component $i$. This formulation evaluates the incremental impact of adding component $i$ to every feasible coalition, adhering to a theoretically rigorous methodology for assessing individual contributions~\cite{shapley1953value}.

\subsection{KernelSHAP}
\label{sec:kernel}

KernelSHAP~\cite{lundberg2017unified} reformulates the Shapley value estimation as a weighted least squares optimization problem, leveraging strategic coalition sampling to efficiently approximate the solution. Formally, the Shapley values $\Phi \in \mathbb{R}^N$  are obtained by solving:
\begin{equation}
\begin{aligned}
    \label{eq:efficiency}
    &\text{min}_{\Phi}\;\mathbb{E}_{p(\mathbf{s})} \Big[ \big( v(\mathbf{s}) - v(\mathbf{0}) - \mathbf{s}^\top \Phi \big)^2 \Big]  \\
    &\quad\: \mathrm{s.t.} \quad\; \mathbf{1}^\top \Phi = v(\mathbf{1}) - v(\mathbf{0}). 
\end{aligned}
\end{equation}
\noindent
The coalition weighting kernel $p(\mathbf{s})$, derived from~\cite{charnes1988extremal}, prioritizes coalitions with very small or large subsets:
\begin{equation}
    p(\mathbf{s}) \propto \frac{(N - 1)(\mathbf{1}^\top \mathbf{s}-1)!(N-\mathbf{1}^\top \mathbf{s}-1)!}{N!} \, ,\label{eq:shapley-kernel}
\end{equation}
\noindent
where $0 < \mathbf{1}^\top \mathbf{s} < N$. 
Theoretical analyses \cite{lundberg2017unified,covert2021improving} demonstrate that this estimator is statistically consistent and unbiased under mild assumptions. However, the computational cost remains prohibitive, as separate optimization problems must be solved for every image $\mathbf{x}$.

\section{Method}

\label{sec:method}

In this section, we propose a dual-branch framework that jointly optimizes the primary task (e.g., image classification) and Shapley value estimation to enhance the interpretability of predictive models. First, the input image $\mathbf{x}$ is partitioned into $N$ patches, each treated as an independent component. The first branch processes the original image and employs a patch-based soft voting for the primary task, where the contribution of each patch is weighted by its predicted Shapley value (i.e., explanations). The second branch processes the masked image to supervise the explanations. The framework overview is shown in Figure~\ref{fig:4}.

\subsection{Patch-Level Decomposition and Representation}
\label{sec:patch process}

Given an input image $\mathbf{x} \in \mathbb{R}^{C \times H \times W}$, we decompose it into $N = H_p \times W_p$ patches with patch size $P$, each treated as a component. These patches are processed through a vision model $\mathbf{\Phi}$ with parameters $\theta$, comprising an encoder $\text{Enc}(\cdot)$ and a prediction head $\text{Head}(\cdot)$. The model directly predicts the Shapley value of each patch as follows:
\begin{align}
    \mathbf{E} &= \text{Enc}(\mathbf{x}) \in \mathbb{R}^{N \times L}, \\
    \mathbf{\Phi}(\mathbf{x}; \theta) &= \text{Head}(\mathbf{E}) \in \mathbb{R}^{N \times K},
\end{align}
where $L$ denotes the feature dimension per patch, and $\mathbf{\Phi}(\mathbf{x}; \theta)$ represents class-specific shapley values for $K$ classes across all $N$ patches.

\subsection{Patch-Based Soft Voting}

In our framework, image-level predictions are then obtained by aggregating predicted Shapley values $\mathbf{\Phi}(\mathbf{x}; \theta)$ for each patch through a soft voting mechanism:
\begin{equation} \label{eq:sum function}
\mathbf{f}(\mathbf{x};\theta) = \sum_{n=1}^{N}\mathbf{\Phi}(\mathbf{x}; \theta)_{n,:}  \in \mathbb{R}^{K}. 
\end{equation}
\noindent
This aggregation mechanism, rooted in cooperative game theory, ensures fair attribution of each patch's marginal contribution to the final outcome of the primary task.  Taking image classification as an example, the objective is then optimized  using a standard cross-entropy loss:
\begin{equation}
\ell_\mathrm{cls} = \mathbb{E}_{p(\mathbf{x})} \text{CrossEntropy}(\mathbf{f}(\mathbf{x}; \theta), y^*),
\end{equation}
\noindent
where $y^*$ denotes the ground truth label of $\mathbf{x}$.

\subsection{Shapley Value Estimation via Self-Supervised Learning}
We present our self-supervised framework for Shapley value estimation using the formulation in Problem~\eqref{eq:efficiency}.
\subsubsection{Mask Sampling} 

The process begins by strategically sampling the binary mask $\mathbf{s}$ from the Shapley kernel defined in Eq.~\ref{eq:shapley-kernel}. Each mask represents a coalition of patches, where $\mathbf{s}_i=1$ indicates the inclusion of patch $i$ in the coalition. 

\subsubsection{Masked Image Processing} 
\label{sec:mask}
Building on the patch decomposition outlined previously, we process each input image $\mathbf{x}$ by applying the sampled mask $\mathbf{s}$ to discard the excluded patches. We employ specialized masking strategies for different architectures: standard zero-masking for CNN, and physical patch removal for ViT similar to the approach proposed in Masked Autoencoders~\cite{he2022masked}. The resulting masked image $\mathbf{x}_{\mathbf{s}}$ is then encoded through:
\begin{align}
    \mathbf{E}_{\mathbf{s}} &= \text{Enc}(\mathbf{x}_{\mathbf{s}}) \in \mathbb{R}^{M \times L}, \\
    \mathbf{\Phi}(\mathbf{x}_{\mathbf{s}}; \theta) &= \text{Head}(\mathbf{E}_{\mathbf{s}} ) \in \mathbb{R}^{M \times K},
\end{align}
where $M$ is the number of included patches. 

\subsubsection{Shapley Loss Formulation}
Following  FastSHAP~\cite{jethani2021fastshap}, we estimate Shapley values through a parameterized function $\mathbf{\Phi}(\mathbf{x}; \theta)$, which simultaneously serves as the prediction mechanism via the aggregation in Eq.~\ref{eq:sum function}. In addition to the classification loss, we optimize $\theta$ through the Shapley loss derived from Problem~\eqref{eq:efficiency}:

\begin{equation}
    \ell_\mathrm{shap} = \mathbb{E}_{p(\mathbf{x}), p(y), p(\mathbf{s})} \Big[ \big( v_{y}(\mathbf{s}) - v_{y}(\mathbf{0}) - \mathbf{s}^\top \mathbf{\Phi}(\mathbf{x};\theta)_{:,y}\big)^2 \Big],
\end{equation}
\noindent 
where $p(y)$  denotes uniform class sampling, and $\mathbf{\Phi}(\mathbf{x};\theta)_{:,y}$ denotes class-specific Shapley value for class $y$.

We implement patch masking through either zero-masking or physical removal. This methodology establishes the baseline value $v(\mathbf{0})$ as equivalent to the model's empty-input prediction. By standardizing $v(\mathbf{0})=0$ to streamline optimization, the constraint in Problem~\eqref{eq:efficiency} simplifies to: 

\begin{equation}
    \mathbf{1}^\top \mathbf{\Phi}(\mathbf{x};\theta)_{:,y}  = v_{y}(\mathbf{1}),
\end{equation}
and the constraint is inherently satisfied by the aggregation mechanism defined in Eq.~\ref{eq:sum function} and the value function formulation in Eq.~\ref{eq:value function}. The Shapley loss is thus simplified as a self-supervised objective:
\begin{equation}
    \ell_\mathrm{shap} = \mathbb{E} \Big[( \sum_{m=1}^{M}\mathbf{\Phi}(\mathbf{x}_{\mathbf{s}}; \theta)_{m,y} - \mathbf{s}^\top \mathbf{\Phi}(\mathbf{x};\theta)_{:,y} )^2  \Big], 
\end{equation}
where $\sum_{m=1}^{M}\mathbf{\Phi}(\mathbf{x}_{\mathbf{s}}; \theta)_{m,y}$ is the predictive value for class $y$ given masked input $\mathbf{x}_{\mathbf{s}}$, which is equal to $v_y(\mathbf{s})$.

\subsection{Training Procedure}

The total loss is a combination of the primary objective (e.g., classification loss)  and Shapley loss as:
\begin{equation}\label{eq:total_loss}
    \ell = \ell_\mathrm{cls} + \lambda \cdot \ell_\mathrm{shap} ,
\end{equation}
\noindent

\noindent
where $\lambda$ serves as a trade-off hyperparameter. The pseudo-code of the training procedure is provided in the Appendix.

\section{Experiments}
\label{sec:exp}
This section presents a comprehensive evaluation of our proposed self-explaining model through multiple experiments across diverse datasets and applications. We begin with model implementations, followed by extensive comparisons with state-of-the-art methods. Code is available at https://github.com/kalofan/SENNs-via-SVO.

\subsection{Method Implementations}
\label{sec:implement}

We evaluate our framework using two standard architectures, ViT-B and ResNet-50, with an input image size of $3 \times 224 \times 224$. While preserving the core functionality of each model, we introduce specific adaptations to achieve patch decomposition and spatial feature representation critical for interpretable Shapley value estimation. 

\noindent\textbf{Feature Extraction}. For ViT, we remove the class token \texttt{[cls]} and retain all the $N$ patch tokens, generating patch feature representations with the size of $N \times L$. For CNN architectures, we preserve the inherent structures while reshaping outputs to match the ViT format. Specifically, a standard ResNet-50 encoder produces feature maps of dimensions $L \times \sqrt{N} \times \sqrt{N}$, which we reshape to $N \times L$ patch representations:
$\mathbf{E} = \text{Reshape}(\mathrm{Enc}_{\mathrm{CNN}}(\mathbf{x}))$.

\noindent\textbf{Masked Feature Extraction}. Given masks $\mathbf{s}$, we retain only $M$ unmasked patches and remove masked patches along with \texttt{[cls]} for ViT, thereby extracting patch representations with the size of $M \times L$. For CNN models, direct removal of masked patches is infeasible. Instead, we first generate masked input $\mathbf{x}_{\mathbf{s}}$ by zeroing out masked patches. $\mathbf{x}_{\mathbf{s}}$ is then processed through the CNN encoder, followed by a Gather operation to select patch features exclusively from unmasked positions:
\begin{equation}
\mathbf{E}_{\mathbf{s}} = \text{Gather}(\text{Reshape}(\mathrm{Enc}_{\mathrm{CNN}}(\mathbf{x}_{\mathbf{s}})), \mathbf{s})
\end{equation}

\noindent\textbf{Hyperparameters}. For ViT-B, we employ a patch size of $P=16$, producing $N=196$ with a feature dimension $L=768$. For ResNet-50, we modify the stride parameter from $2$ to $1$ in the initial residual block's convolutional layer in stage $4$, generating feature maps of dimensions $L \times \sqrt{N} \times \sqrt{N} = 2048 \times 14 \times 14$, which are reshaped to $N \times L = 196 \times 2048$. To accelerate convergence, all backbones are initialized with ImageNet pre-trained weights and fine-tuned by our method, underscoring our method’s compatibility with standard architectures. Additional training details and time costs are provided in the Appendix.

\begin{table}
\centering
\small
\begin{tabular}{l|ccc}
\toprule 
Model &  Acc@1 & \#Params &\#FLOPs \\
\midrule 
 BagNet & 63.5 & 18M & 16.4B \\
 HarsanyiNet & 67.1 & 45M &  6.1B\\
 \midrule
ViT-B & 75.0  & 86M & 17.6B\\
B-cos-V & 74.4& 86M & 17.6B \\
B-cosified-V & 75.3 & 86M & 17.8B \\
 Ours-V & \textbf{76.8}  & 86M & 17.6B\\
\midrule
ResNet-50  & 75.3 & 26M & 4.1B \\
ProtoPNet & 72.3 & 29M & 4.4B\\
$\mathcal{X}$-DNNs  & 72.9 & 26M & 4.1B \\ 
B-cos-R & 75.9& 26M & 4.3B \\
B-cosified-R & 76.5 & 26M & 4.3B \\
 Ours-R & \textbf{76.3}  & 26M & 6.6B\\
\bottomrule
\end{tabular}
\caption{Performance comparison of different models on ImageNet classification. The table is organized into three sections: custom CNNs (top), ViT-B-based models ("-V", middle), and ResNet-50-based models ("-R", bottom).}     
\label{tab:1} 
\end{table}

\begin{table}[t]
\centering
\small
\begin{tabular}{ll|ccc}
\toprule 
Backbone&XAI Method& AOPC$\uparrow$ & LOdds$\downarrow$ & SaCo$\uparrow$ \\
\midrule
 ViT-B&Grad-CAM &0.547& -4.179& 0.1187\\
 ViT-B&Rollout & 0.671& -5.186& 0.2887\\
 ViT-B&GAME &0.707&-5.391&0.4353\\
 ViT-B&CheferLRP &0.715&-5.587&0.4411\\
 ViT-B&ATTCAT &0.647&-4.387&0.3454\\
 ViT-B&ViTShapley &0.676&-5.056&0.4095\\
 ViT-B&TokenTM &0.755&-5.668&0.4827\\
\midrule
 ResNet-50&IG& 0.643& -4.689& 0.2929\\
 ResNet-50&Grad-CAM& 0.604& -4.275& 0.2182\\
 ResNet-50&FastSHAP&0.683&-4.976&0.3197\\
 ResNet-50&IDGI&0.706&-5.225&0.4057\\
 ResNet-50&CAE &0.601&-4.107&0.2224\\
\midrule
ProtoPNet&ProtoPNet & 0.577& -4.075& 0.2053\\
BagNet&BagNet & 0.794& -5.864& 0.5124\\
$\mathcal{X}$-DNNs&$\mathcal{X}$-DNNs  & 0.636& -4.678& 0.2703\\  
B-cos-V&B-cos & 0.714& -5.474& 0.4334\\
B-cos-R&B-cos& 0.717& -5.486& 0.4175\\
HarsanyiNet&HarsanyiNet & 0.798& -5.903& 0.5147\\
B-cosified-V&B-cosified& 0.712& -5.473& 0.4323\\
B-cosified-R&B-cosified& 0.718& -5.489& 0.4207\\
Ours-V&Ours &  0.803&  -5.987&  \textbf{0.5196}\\
Ours-R&Ours &  \textbf{0.806}&  \textbf{-6.036}&  0.5193\\
\bottomrule
\end{tabular}
\caption{Comparison of explanation performance on the ImageNet dataset. The upper and middle sections report post-hoc methods applied to vanilla ViT-B and ResNet-50 backbones, respectively, while the lower section shows results for various SENNs. The “-V” and “-R” suffixes indicate ViT and ResNet backbones, respectively.}
\label{tab:2}
\end{table}

\begin{table}[t]
\centering
\small
\begin{tabular}{ll|ccc}
\toprule 
Backbone&XAI Method & Pix. Acc.$\uparrow$ & mAP$\uparrow$ & mIoU$\uparrow$ \\
\midrule
ViT-B&Grad-CAM&66.83&77.89 &45.14 \\
ViT-B&Rollout & 58.18&73.63 &39.25 \\
ViT-B&GAME &78.17&85.15&60.18\\
ViT-B&CheferLRP &79.89&85.63&60.97\\
ViT-B&ATTCAT &41.18&47.89&28.05\\
ViT-B&ViTShapley &77.47&80.92&58.18\\
ViT-B&TokenTM &81.43&85.17&64.77\\
\midrule
ResNet-50&IG&64.34&68.66&40.78\\
ResNet-50&Grad-CAM& 68.92&79.51 & 45.34 \\
ResNet-50&FastSHAP&73.92&79.53&56.67\\
ResNet-50&IDGI&76.42&79.19&58.82\\
ResNet-50&CAE &61.87&73.73&40.63\\
\midrule
ProtoPNet&ProtoPNet & 59.76& 64.48& 40.82\\
BagNet&BagNet & 82.76& 86.72& 66.77\\
$\mathcal{X}$-DNNs  &$\mathcal{X}$-DNNs  & 64.21&72.94 & 49.01\\  
B-cos-V&B-cos&75.56& 80.27& 60.83\\
B-cos-R&B-cos&77.92& 82.16& 61.32\\
HarsanyiNet &HarsanyiNet & 83.16&86.97 & 66.03\\
B-cosified-V&B-cosified& 75.57& 80.86& 60.43\\
B-cosified-R&B-cosified& 77.91& 81.97& 61.31\\
Ours-V & Ours &   83.79& 86.98 &  66.38\\
Ours-R &Ours & \textbf{85.78}&  \textbf{87.67}&  \textbf{66.96}\\
\bottomrule
\end{tabular}
\caption{Quantitative comparison of segmentation performance on the ImageNet-Segmentation dataset. Explanations from various methods are converted into binary segmentation masks and evaluated using Pix. Acc., mAP and mIoU. The upper and center sections present results for applying post-hoc explanation methods to vanilla ViT-B and ResNet-50 backbones, respectively, while the lower section shows results for various SENNs.}
\label{tab:3}
\end{table}

\subsection{Vision Models}

\subsubsection{Prediction Performance}
Table~\ref{tab:1} presents a comparative analysis of our method against pre-trained ResNet-50, ViT-B, and state-of-the-art SENNs in terms of classification performance on ImageNet~\cite{russakovsky2015imagenet}. Evaluated SENNs include ProtoPNet~\cite{chen2019looks}, BagNet~\cite{brendel2019approximating}, $\mathcal{X}$-DNNs~\cite{hesse2021fast}, B-cos~\cite{bohle2022bcos}, HarsanyiNet~\cite{chen2023harsanyinet} and B-cosified~\cite{arya2025bcosification}. Metrics include  top-1 accuracy (Acc@$1$), number of model parameters, and computational complexity (FLOPs). BagNet and HarsanyiNet, shown in the upper section of the table, employ custom CNN models. The models in the center section are based on ViT-B architectures, whereas those in the lower section primarily utilize ResNet-50 variants.

Our ViT-B implementation achieves $76.8\%$ top-1 accuracy, surpassing the vanilla ViT-B ($75.0\%$) by $1.8\%$ while retaining identical parameters and FLOPs. This demonstrates that our interpretability framework enhances rather than compromises classification performance. Similarly, when applied to the ResNet-50, our method achieves $76.3\%$ accuracy, outperforming the vanilla ResNet-50 ($75.3\%$). This improvement is particularly significant given that existing interpretable models such as ProtoPNet and $\mathcal{X}$-DNNs typically exhibit substantial performance degradation. These results validate that integrating Shapley value estimation during training not only maintains but enhances classification performance, while introducing a minor to no increase in computational complexity.

\subsubsection{Explanation Faithfulness}

To evaluate the explanation faithfulness, we employ multiple complementary metrics on ImageNet: AOPC (Area Over the Perturbation Curve)~\cite{chen2020generating}, LOdds (Log-odds)~\cite{qiang2022attcat}, and SaCo (Salience-guided Faithfulness Coefficient)~\cite{wu2024faithfulness}. AOPC measures the average drop in model confidence when perturbing the most salient pixels, and a higher AOPC is better. LOdds evaluates if the most important pixels are sufficient for the prediction by measuring the log-odds after perturbation, and a lower LOdds is better. SaCo evaluates faithfulness by correlating pixel saliency scores with their actual impact on the model's prediction, and a higher SaCo indicates a more faithful explanation. In addition, the number of pixel subsets for SaCo computation is $10$.

We compare our method with several SOTA explaining methods (XAI), including post-hoc methods and SENNs. For post-hoc explanation on the vanilla ViT-B backbone, we evaluate Grad-CAM~\cite{Selvaraju2017gradcam}, Rollout~\cite{abnar2020quantifying}, GAME~\cite{chefer2021generic}, CheferLRP~\cite{chefer2021transformer}, ATTCAT~\cite{qiang2022attcat}, ViTShapley~\cite{covert2023learning}, and TokenTM~\cite{wu2024token}. For vanilla ResNet-50 backbone, we evaluate IG~\cite{sundararajan2017axiomatic}, Grad-CAM~\cite{Selvaraju2017gradcam}, FastSHAP~\cite{jethani2021fastshap}, IDGI~\cite{yang2023idgi} and CAE~\cite{xie2024accurate}. We also compare our self-explaining framework with representative SENNs, including ProtoPNet, BagNet, $\mathcal{X}$-DNN, B-cos Network, HarsanyiNet, and B-cosified Network.

As shown in Table~\ref{tab:2}, our method achieves superior performance on ImageNet, demonstrating that our explanations faithfully represent the model's decision-making processes. Our framework’s intrinsic explanations significantly outperform post-hoc methods on ImageNet. Notably, our approach achieves consistent interpretability performance across different backbones. Though our ResNet-50 variant shows marginally better overall performance, the ViT-B implementation achieves competitive results across all metrics. Among existing SENNs, BagNet and HarsanyiNet demonstrate competitive interpretability, but suffer severe degradation of classification performance (Table~\ref{tab:1}). In contrast, our method maintains high performance in both classification and explanation, requiring minor architectural modifications. 

\begin{figure*}
\centering
\includegraphics[width=0.95\textwidth]{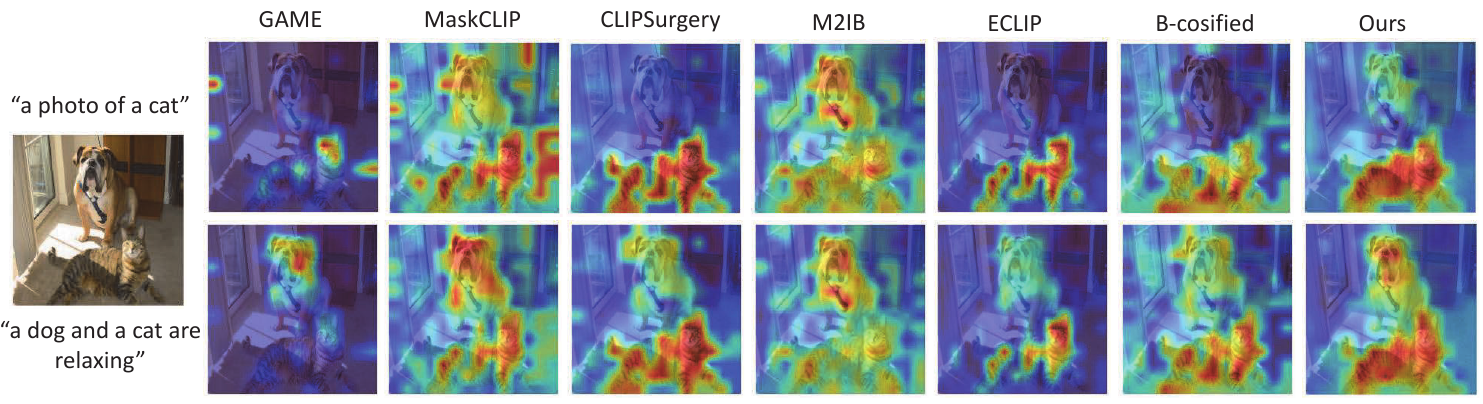}
\caption{Visualization of different explanation methods for image-text matching. The left column shows input texts matched with the image, and subsequent columns display corresponding explanations generated by various methods.}
\label{fig:itm-vis}
\end{figure*}

\begin{table}
\centering
\small 
\begin{tabular}{llcccc} 
\toprule
\multirow{2}{*}{Dataset}&\multirow{2}{*}{Method}& \multicolumn{2}{c}{T2I} & \multicolumn{2}{c}{I2T} \\
\cmidrule(lr){3-4} \cmidrule(lr){5-6} 
&& R@1 & R@5 & R@1 & R@5  \\
\midrule
&CLIP & 42.2& 66.9& 58.7& 80.7\\
MS COCO&B-cosified & 41.9& 66.7& 59.1& 81.1\\
&Ours & 42.1& 67.3& 58.9& 80.9\\
\midrule
&CLIP & 71.2& 91.0& 85.7& 96.7\\
Flickr30k&B-cosified & 70.8& 90.7& 86.0& 97.1\\
&Ours & 70.8& 91.1& 85.5& 97.1\\
\bottomrule
\end{tabular}
\caption{Comparison of zero-shot retrieval performance on MS COCO and Flickr30k datasets. We report Recall@1 (R@1) and Recall@5 (R@5) for both text-to-image (T2I) and image-to-text (I2T) retrieval tasks.}
\label{tab:retrieval_comparison}
\end{table}

\begin{table}[t]
\centering
\small
\begin{tabular}{lcccc}
\toprule
\multirow{2}{*}{Method} & \multicolumn{2}{c}{Ins. $\uparrow$ on T2I } & \multicolumn{2}{c}{Ins.$\uparrow$ on I2T } \\
\cmidrule(lr){2-3} \cmidrule(lr){4-5}
 & R@1 & R@5 & R@1 & R@5 \\
\midrule
Grad-CAM &0.1027& 0.2216 & 0.1152 & 0.2327\\
Rollout  &0.1294 & 0.2932 & 0.1753 & 0.3503\\
GAME  &0.1537 & 0.3083 & 0.2097 & 0.3735\\
MaskCLIP &0.1423& 0.2953& 0.1891 & 0.3514\\
CLIPSurgery &0.1419& 0.2941&0.1771& 0.3384\\
M2IB &0.1469&0.3004&0.2058&0.3691\\
ECLIP &0.1576&0.3203&0.2056&0.3761\\
B-cosified &0.1558&0.3089&0.2016&0.3622\\
Ours &\textbf{0.1679}&\textbf{0.3360}&\textbf{0.2183}&\textbf{0.3902}\\
\bottomrule
\end{tabular}
\caption{Evaluation of explanation on MS COCO dataset: Ins. on image retrieval and text retrieval tasks.}
\label{tab:coco_ins}
\end{table}

\begin{table}
\centering
\small
\begin{tabular}{lcccc}
\toprule
\multirow{2}{*}{Method} & \multicolumn{2}{c}{Del.$\downarrow$ on T2I } & \multicolumn{2}{c}{Del.$\downarrow$ on I2T } \\
\cmidrule(lr){2-3} \cmidrule(lr){4-5}
 & R@1 & R@5 & R@1 & R@5 \\
\midrule
Grad-CAM &0.1717 & 0.3502& 0.2161 & 0.4008\\
Rollout  &0.1948&0.3946&0.2268&0.4238\\
GAME  &0.1706&0.3552&0.1982&0.3800\\
MaskCLIP &0.1321&0.2841&0.1516&0.2949\\
CLIPSurgery&0.1794&0.3652&0.2381&0.4292\\
M2IB &0.1797&0.3671&0.2057&0.3905\\
ECLIP &0.1246 & 0.2670 & 0.1550&0.2933\\
B-cosified &0.1257 & 0.2633 & 0.1632 & 0.2975\\
Ours &\textbf{0.1175}&\textbf{0.2498}&\textbf{0.1401} &\textbf{0.2821}\\
\bottomrule
\end{tabular}
\caption{Evaluation of explanation on MS COCO dataset: Del. on image retrieval and text retrieval tasks.}
\label{tab:coco_del}
\end{table}

\subsubsection{Localization Ability}
Following previous works~\cite{wu2024token, zhao2024gradient,chefer2021transformer}, we perform segmentation on the ImageNet-Segmentation~\cite{gao2022large} dataset, treating explanation heatmaps as preliminary semantic signals and using a predefined threshold to produce binary segmentation maps. Based on the ground-truth maps, we evaluate the performance on three metrics: pixel-wise accuracy (Pix. Acc.), mean average precision (mAP), and mean intersection over union (mIoU).

The quantitative results, presented in Table~\ref{tab:3}, unequivocally demonstrate the superiority of our framework in generating spatially precise explanations that are directly applicable to semantic segmentation. Our method consistently outperforms a wide array of both post-hoc and self-explaining methods across two distinct architectures. The superior performance in this segmentation task validates that our method learns explanations that are not only faithful to the model's classification logic but also possess high spatial fidelity. The learned Shapley values effectively highlight the complete extent of the object, making them a valuable and reliable signal for dense prediction tasks.

\subsection{Contrastive Language-Image Pre-Trained Models}

To further demonstrate the generalization capabilities and multi-modal applicability of our proposed self-explaining framework, we extend its evaluation to the image-text matching task. Following ECLIP~\cite{zhao2024gradient}, we adopt EVA-CLIP (ViT-B/16)~\cite{sun2023eva} as the baseline model. Modifications are applied exclusively to the vision encoder to incorporate Shapley value estimation, while the text encoder remains unchanged. Mirroring the implementation in the classification framework in Figure~\ref{fig:4}, the dual-branch framework processes both unmasked and masked images through separate pathways. Key adaptations include: 1) replace the classification head with a cosine similarity measurement between patch and text embeddings, and 2) use the global sum-pooling to aggregate patch-wise similarities into an image-text similarity, which is capable of the standard  CLIP contrastive learning.

Our training objective comprises the standard CLIP contrastive loss and the Shapley loss, enabling the model to generate explanations for image-text alignment through the identification of salient patches corresponding to textual descriptions, while maintaining cross-modal alignment capabilities. For a fair comparison, both our method and B-cosified (the competing self-explaining approach) are fine-tuned on the prevalent large-scale dataset CC3M~\cite{sharma2018conceptual} dataset from EVA-CLIP pre-trained weights. The training details are in the Appendix.

\subsubsection{Prediction Performance}
We evaluate zero-shot retrieval performance on MS COCO~\cite{lin2014microsoft} and Flickr30k~\cite{plummer2015flickr30k} using Recall@1 (R@1) and Recall@5 (R@5). As presented in Table~\ref{tab:retrieval_comparison}, our method achieves zero-shot retrieval performance that is highly competitive with the vanilla CLIP baseline across both MS COCO and Flickr30k datasets. This confirms that the integration of the Shapley-based optimization preserves primary task capability.

\subsubsection{Explanation Faithfulness}
Following ECLIP~\cite{zhao2024gradient}, we evaluate explanation faithfulness using rigorous perturbation tests on MS COCO: Insertion (Ins.) and Deletion (Del.) metrics. Our method is compared against state-of-the-art CLIP explanation approaches, including Grad-CAM, Rollout, GAME, MaskCLIP~\cite{zhou2022extract}, CLIPSurgery~\cite{li2023clip}, M2IB~\cite{wang2023visual}, ECLIP~\cite{zhao2024gradient}, and B-cosified. As Tables \ref{tab:coco_ins} and \ref{tab:coco_del} demonstrate, our approach achieves optimal scores on both metrics, confirming that identified pixels are indeed critical to the model's decision-making process. Visually, Figure~\ref{fig:itm-vis} illustrates the explanation faithfulness of our method, showing precise alignment between highlighted patches and textual descriptions.

Further experiments, including an ablation study on the trade-off hyperparameter $\lambda$, and an analysis of Shapley value estimation errors are presented in the Appendix.

\section{Conclusion}
\label{sec:conclusion}

In this paper, we introduce a novel self-explaining framework that integrates Shapley value estimation into the training process. Extensive experiments demonstrate that our method not only enhances primary task performance but also provides faithful explanations. These results substantiate that integrating Shapley value estimation through multitask learning offers an optimal balance between model performance and interpretability. 

\section{Acknowledgments}
This work was supported by the Early Career Scheme (No. CityU 21219323) and the General Research Fund (No. CityU 11220324) of the University Grants Committee (UGC), the NSFC Young Scientists Fund (No. 9240127), and the Donation for Research Projects (No. 9229164 and No. 9220187).

\bibliography{aaai2026}

@inproceedings{he2016deep,
title={Deep residual learning for image recognition},
author={He, Kaiming and Zhang, Xiangyu and Ren, Shaoqing and Sun, Jian},
booktitle={IEEE Conference on Computer Vision and Pattern Recognition},
pages={770--778},
year={2016}}

@inproceedings{he2022masked,
  title={Masked autoencoders are scalable vision learners},
  author={He, Kaiming and Chen, Xinlei and Xie, Saining and Li, Yanghao and Doll{\'a}r, Piotr and Girshick, Ross},
  booktitle={IEEE Conference on Computer Vision and Pattern Recognition},
  pages={16000--16009},
  year={2022}
}

@InProceedings{Selvaraju2017gradcam,
author = {Selvaraju, Ramprasaath R. and Cogswell, Michael and Das, Abhishek and Vedantam, Ramakrishna and Parikh, Devi and Batra, Dhruv},
title = {{Grad-CAM}: {Visual} Explanations From Deep Networks via Gradient-Based Localization},
booktitle = {IEEE International Conference on Computer Vision},
pages = {618--626},
year = {2017}
}

@phdthesis{nauta2023explainable,
  title={Explainable {AI} and Interpretable Computer Vision: {From} Oversight to Insight},
  author={Nauta, Meike},
  type = {PhD Thesis},
  school = {University of Twente},
  year={2023}
}

@inproceedings{nauta2021neural,
  title={Neural prototype trees for interpretable fine-grained image recognition},
  author={Nauta, Meike and Van Bree, Ron and Seifert, Christin},
  booktitle={IEEE Conference on Computer Vision and Pattern Recognition},
  pages={14933--14943},
  year={2021}
}

@inproceedings{chen2019looks,
  title={This looks like that: {Deep} learning for interpretable image recognition},
  author={Chen, Chaofan and Li, Oscar and Tao, Daniel and Barnett, Alina and Rudin, Cynthia and Su, Jonathan K.},
  booktitle={Advances in Neural Information Processing Systems},
  year={2019}
}

@inproceedings{ribeiro2016should,
  title={"{Why} should {I} trust you?" {Explaining} the predictions of any classifier},
  author={Ribeiro, Marco Tulio and Singh, Sameer and Guestrin, Carlos},
  booktitle={ACM International Conference on Knowledge Discovery and Data Mining},
  pages={1135--1144},
  year={2016}
}

@inproceedings{sundararajan2017axiomatic,
  title={Axiomatic attribution for deep networks},
  author={Sundararajan, Mukund and Taly, Ankur and Yan, Qiqi},
  booktitle={International Conference on Machine Learning},
  pages={3319--3328},
  year={2017}
}

@article{borys2023explainable,
  title={{Explainable AI} in Medical Imaging: {An} overview for clinical practitioners---Saliency-based {XAI} approaches},
  author={Borys, Katarzyna and Schmitt, Yasmin Alyssa and Nauta, Meike and Seifert, Christin and Kr{\"a}mer, Nicole and Friedrich, Christoph M. and Nensa, Felix},
  journal={European Journal of Radiology},
  volume={162},
  number={110787},
  year={2023}
}

@inproceedings{covert2021improving,
  title={{Improving kernelSHAP}: {Practical} shapley value estimation using linear regression},
  author={Covert, Ian and Lee, Su-In},
  booktitle={International Conference on Artificial Intelligence and Statistics},
  pages={3457--3465},
  year={2021}
}

@article{kindermans2019reliability,
  title={The (un) reliability of saliency methods},
  author={Kindermans, Pieter-Jan and Hooker, Sara and Adebayo, Julius and Alber, Maximilian and Sch{\"u}tt, Kristof T and D{\"a}hne, Sven and Erhan, Dumitru and Kim, Been},
  journal={Explainable AI: Interpreting, Explaining and Visualizing Deep Learning},
  volume={11700},
  pages={267--280},
  year={2019}
}

@article{petsiuk2018rise,
  title={Rise: {Randomized} input sampling for explanation of black-box models},
  author={Petsiuk, Vitali and Das, Abir and Saenko, Kate},
  journal={arXiv preprint arXiv:1806.07421},
  year={2018}
}

@inproceedings{lundberg2017unified,
title = {A Unified Approach to Interpreting Model Predictions},
 author = {Lundberg, Scott M. and Lee, Su-In},
 booktitle = {Advances in Neural Information Processing Systems},
 year = {2017}
}

@inproceedings{jethani2021fastshap,
  title={Fastshap: {Real-time} shapley value estimation},
  author={Jethani, Neil and Sudarshan, Mukund and Covert, Ian Connick and Lee, Su-In and Ranganath, Rajesh},
  booktitle={International Conference on Learning Representations},
  year={2022}
}

@article{abnar2020quantifying,
  title={Quantifying attention flow in transformers},
  author={Abnar, Samira and Zuidema, Willem},
  journal={arXiv preprint arXiv:2005.00928},
  year={2020}
}

@inproceedings{dosovitskiy2020image,
  title={An image is worth 16x16 words: {Transformers} for image recognition at scale},
  author={Dosovitskiy, Alexey and Beyer, Lucas and Kolesnikov, Alexander and Weissenborn, Dirk and Zhai, Xiaohua and Unterthiner, Thomas and Dehghani, Mostafa and Minderer, Matthias and Heigold, Georg and Gelly, Sylvain and others},
  booktitle={International Conference on Learning Representations},
  year={2020}
}

@inproceedings{fong2019understanding,
  title={Understanding deep networks via extremal perturbations and smooth masks},
  author={Fong, Ruth and Patrick, Mandela and Vedaldi, Andrea},
  booktitle={IEEE International Conference on Computer Vision},
  pages={2950--2958},
  year={2019}
}

@inbook{shapley1953value,
title = {A Value for n-Person Games},
author = {Shapley, Lloyd S.},
booktitle = {Contributions to the Theory of Games, Volume II},
publisher = {Princeton University Press},
address = {Princeton},
pages = {307--318},
year = {1953}
}

@article{russakovsky2015imagenet,
  title={{ImageNet} large scale visual recognition challenge},
  author={Russakovsky, Olga and Deng, Jia and Su, Hao and Krause, Jonathan and Satheesh, Sanjeev and Ma, Sean and Huang, Zhiheng and Karpathy, Andrej and Khosla, Aditya and Bernstein, Michael and  Bergs, Alexander C. and Li, Fei-Fei},
  journal={International Journal of Computer Vision},
  volume={115},
  pages={211--252},
  year={2015}
}

@inproceedings{redmon2016you,
  title={You only look once: {Unified}, real-time object detection},
  author={Redmon, Joseph and Divvala, Santosh and Girshick, Ross and Farhadi, Ali},
  booktitle={IEEE Conference on Computer Vision and Pattern Recognition},
  pages={779--788},
  year={2016}
}

@inproceedings{vinyals2015show,
  title={Show and tell: {A} neural image caption generator},
  author={Vinyals, Oriol and Toshev, Alexander and Bengio, Samy and Erhan, Dumitru},
  booktitle={IEEE Conference on Computer Vision and Pattern Recognition},
  pages={3156--3164},
  year={2015}
}

@inproceedings{antol2015vqa,
  title={{VQA}: {Visual} question answering},
  author={Antol, Stanislaw and Agrawal, Aishwarya and Lu, Jiasen and Mitchell, Margaret and Batra, Dhruv and Zitnick, C. Lawrence and Parikh, Devi},
  booktitle={IEEE International Conference on Computer Vision},
  pages={2425--2433},
  year={2015}
}

@book{molnar2020interpretable,
  title={Interpretable machine learning},
  author={Molnar, Christoph},
  year={2020},
  publisher={Lulu. com}
}

@article{hesse2024benchmarking,
  title={Benchmarking the Attribution Quality of Vision Models},
  author={Hesse, Robin and Schaub-Meyer, Simone and Roth, Stefan},
  journal={arXiv preprint arXiv:2407.11910},
  year={2024}
}

@inproceedings{bohle2022bcos,
  title={B-cos networks: {Alignment} is all we need for interpretability},
  author={B{\"o}hle, Moritz and Fritz, Mario and Schiele, Bernt},
  booktitle={IEEE Conference on Computer Vision and Pattern Recognition},
  pages={10329--10338},
  year={2022}
}

@article{brendel2019approximating,
  title={Approximating {CNNs} with bag-of-local-features models works surprisingly well on imagenet},
  author={Brendel, Wieland and Bethge, Matthias},
  journal={arXiv preprint arXiv:1904.00760},
  year={2019}
}

@article{hesse2021fast,
  title={Fast axiomatic attribution for neural networks},
  author={Hesse, Robin and Schaub-Meyer, Simone and Roth, Stefan},
  journal={Advances in Neural Information Processing Systems},
  pages={19513--19524},
  year={2021}
}

@article{wang2021shapley,
  title={Shapley explanation networks},
  author={Wang, Rui and Wang, Xiaoqian and Inouye, David I.},
  journal={arXiv preprint arXiv:2104.02297},
  year={2021}
}

@article{chen2023harsanyinet,
  title={Harsanyinet: Computing accurate shapley values in a single forward propagation},
  author={Chen, Lu and Lou, Siyu and Zhang, Keyan and Huang, Jin and Zhang, Quanshi},
  journal={arXiv preprint arXiv:2304.01811},
  year={2023}
}

@article{castro2009polynomial,
  title={Polynomial calculation of the Shapley value based on sampling},
  author={Castro, Javier and G{\'o}mez, Daniel and Tejada, Juan},
  journal={Computers \& Operations Research},
  volume={36},
  number={5},
  pages={1726--1730},
  year={2009}
}

@article{strumbelj2010efficient,
  title={An efficient explanation of individual classifications using game theory},
  author={Strumbelj, Erik and Kononenko, Igor},
  journal={The Journal of Machine Learning Research},
  volume={11},
  pages={1--18},
  year={2010}
}

@article{mitchell2022sampling,
  title={Sampling permutations for shapley value estimation},
  author={Mitchell, Rory and Cooper, Joshua and Frank, Eibe and Holmes, Geoffrey},
  journal={Journal of Machine Learning Research},
  volume={23},
  number={43},
  pages={1--46},
  year={2022}
}

@incollection{charnes1988extremal,
  title={Extremal principle solutions of games in characteristic function form: core, {C}hebychev and {S}hapley value generalizations},
  author={Charnes, A and Golany, B and Keane, M and Rousseau, J},
  booktitle={Econometrics of Planning and Efficiency},
  pages={123--133},
  year={1988},
  publisher={Springer}
}

@inproceedings{arya2025bcosification,
  title={{B-cosification}: {Transforming} deep neural networks to be inherently interpretable},
  author={Arya, Shreyash and Rao, Sukrut and B{\"o}hle, Moritz and Schiele, Bernt},
  booktitle={Advances in Neural Information Processing Systems},
  pages={62756--62786},
  year={2024}
}

@inproceedings{adebayo2018sanity,
  title={Sanity checks for saliency maps},
  author={Adebayo, Julius and Gilmer, Justin and Muelly, Michael and Goodfellow, Ian and Hardt, Moritz and Kim, Been},
  booktitle={Advances in Neural Information Processing Systems},
  year={2018}
}

@article{yang2019benchmarking,
  title={Benchmarking attribution methods with relative feature importance},
  author={Yang, Mengjiao and Kim, Been},
  journal={arXiv preprint arXiv:1907.09701},
  year={2019}
}

@inproceedings{covert2023learning,
  title={Learning to estimate shapley values with {Vision} {Transformers}},
  author={Covert, Ian and Kim, Chanwoo and Lee, Su-In},
  booktitle={International Conference on Learning Representations},
  year={2023}
}

@inproceedings{zhao2024gradient,
  title={Gradient-based Visual Explanation for CLIP},
  author={Zhao, Chenyang and Wang, Kun and Zeng, Xingyu and Zhao, Rui and Chan, B. Antoni},
  booktitle={International Conference on Machine Learning},
  year = {2024}
}

@inproceedings{chefer2021generic,
  title={Generic attention-model explainability for interpreting bi-modal and encoder-decoder transformers},
  author={Chefer, Hila and Gur, Shir and Wolf, Lior},
  booktitle={International Conference on Computer Vision},
  pages={397--406},
  year={2021}
}

@inproceedings{lin2014microsoft,
  title={Microsoft {COCO}: Common objects in context},
  author={Lin, Tsung-Yi and Maire, Michael and Belongie, Serge and Hays, James and Perona, Pietro and Ramanan, Deva and Doll{\'a}r, Piotr and Zitnick, C Lawrence},
  booktitle={European Conference on Computer Vision},
  pages={740--755},
  year={2014},
}

@inproceedings{yang2023idgi,
  title={IDGI: {A} framework to eliminate explanation noise from integrated gradients},
  author={Yang, Ruo and Wang, Binghui and Bilgic, Mustafa},
  booktitle={IEEE Conference on Computer Vision and Pattern Recognition},
  pages={23725--23734},
  year={2023}
}

@inproceedings{wu2024token,
  title={Token transformation matters: Towards faithful post-hoc explanation for vision transformer},
  author={Wu, Junyi and Duan, Bin and Kang, Weitai and Tang, Hao and Yan, Yan},
  booktitle={IEEE Conference on Computer Vision and Pattern Recognition},
  pages={10926--10935},
  year={2024}
}

@inproceedings{qiang2022attcat,
  title={Attcat: Explaining transformers via attentive class activation tokens},
  author={Qiang, Yao and Pan, Deng and Li, Chengyin and Li, Xin and Jang, Rhongho and Zhu, Dongxiao},
  booktitle={Advances in neural information processing systems},
  pages={5052--5064},
  year={2022}
}

@inproceedings{xie2024accurate,
  title={Accurate explanation model for image classifiers using class association embedding},
  author={Xie, Ruitao and Chen, Jingbang and Jiang, Limai and Xiao, Rui and Pan, Yi and Cai, Yunpeng},
  booktitle={IEEE International Conference on Data Engineering},
  pages={2271--2284},
  year={2024}
}

@article{bass2022icam,
  title={{ICAM}-reg: Interpretable classification and regression with feature attribution for mapping neurological phenotypes in individual scans},
  author={Bass, Cher and Da Silva, Mariana and Sudre, Carole and Williams, Logan ZJ and Sousa, Helena S and Tudosiu, Petru-Daniel and Alfaro-Almagro, Fidel and Fitzgibbon, Sean P and Glasser, Matthew F and Smith, Stephen M and others},
  journal={IEEE Transactions on Medical Imaging},
  volume={42},
  number={4},
  pages={959--970},
  year={2022}
}

@article{li2023negative,
  title={Negative flux aggregation to estimate feature attributions},
  author={Li, Xin and Pan, Deng and Li, Chengyin and Qiang, Yao and Zhu, Dongxiao},
  journal={arXiv preprint arXiv:2301.06989},
  year={2023}
}

@inproceedings{nauta2023pip,
  title={Pip-net: Patch-based intuitive prototypes for interpretable image classification},
  author={Nauta, Meike and Schl{\"o}tterer, J{\"o}rg and Van Keulen, Maurice and Seifert, Christin},
  booktitle={IEEE Conference on Computer Vision and Pattern Recognition},
  pages={2744--2753},
  year={2023}
}

@article{de2024patch,
  title={Patch-based Intuitive Multimodal Prototypes Network (PIMPNet) for Alzheimer's Disease classification},
  author={De Santi, Lisa Anita and Schl{\"o}tterer, J{\"o}rg and Nauta, Meike and Positano, Vincenzo and Seifert, Christin},
  journal={arXiv preprint arXiv:2407.14277},
  year={2024}
}

@article{chen2020generating,
  title={Generating hierarchical explanations on text classification via feature interaction detection},
  author={Chen, Hanjie and Zheng, Guangtao and Ji, Yangfeng},
  journal={arXiv preprint arXiv:2004.02015},
  year={2020}
}

@inproceedings{wu2024faithfulness,
  title={On the faithfulness of vision transformer explanations},
  author={Wu, Junyi and Kang, Weitai and Tang, Hao and Hong, Yuan and Yan, Yan},
  booktitle={IEEE Conference on Computer Vision and Pattern Recognition},
  pages={10936--10945},
  year={2024}
}

@article{gao2022large,
  title={Large-scale unsupervised semantic segmentation},
  author={Gao, Shanghua and Li, Zhong-Yu and Yang, Ming-Hsuan and Cheng, Ming-Ming and Han, Junwei and Torr, Philip},
  journal={IEEE Transactions on Pattern Analysis and Machine Intelligence},
  volume={45},
  number={6},
  pages={7457--7476},
  year={2022},
  publisher={IEEE}
}

@inproceedings{chefer2021transformer,
  title={Transformer interpretability beyond attention visualization},
  author={Chefer, Hila and Gur, Shir and Wolf, Lior},
  booktitle={IEEE Conference on Computer Vision and Pattern Recognition},
  pages={782--791},
  year={2021}
}

@inproceedings{sharma2018conceptual,
  title={Conceptual captions: A cleaned, hypernymed, image alt-text dataset for automatic image captioning},
  author={Sharma, Piyush and Ding, Nan and Goodman, Sebastian and Soricut, Radu},
  booktitle={Annual Meeting of the Association for Computational Linguistics},
  pages={2556--2565},
  year={2018}
}

@inproceedings{plummer2015flickr30k,
  title={Flickr30k entities: Collecting region-to-phrase correspondences for richer image-to-sentence models},
  author={Plummer, Bryan A and Wang, Liwei and Cervantes, Chris M and Caicedo, Juan C and Hockenmaier, Julia and Lazebnik, Svetlana},
  booktitle={IEEE International Conference on Computer Vision},
  pages={2641--2649},
  year={2015}
}

@article{li2023clip,
  title={Clip surgery for better explainability with enhancement in open-vocabulary tasks},
  author={Li, Yi and Wang, Hualiang and Duan, Yiqun and Li, Xiaomeng},
  journal={arXiv e-prints},
  pages={arXiv--2304},
  year={2023}
}

@inproceedings{zhou2022extract,
  title={Extract free dense labels from clip},
  author={Zhou, Chong and Loy, Chen Change and Dai, Bo},
  booktitle={European Conference on Computer Vision},
  pages={696--712},
  year={2022},
}

@inproceedings{wang2023visual,
  title={Visual explanations of image-text representations via multi-modal information bottleneck attribution},
  author={Wang, Ying and Rudner, Tim GJ and Wilson, Andrew G},
  booktitle={Advances in Neural Information Processing Systems},
  pages={16009--16027},
  year={2023}
}

@article{sun2023eva,
  title={Eva-clip: Improved training techniques for clip at scale},
  author={Sun, Quan and Fang, Yuxin and Wu, Ledell and Wang, Xinlong and Cao, Yue},
  journal={arXiv preprint arXiv:2303.15389},
  year={2023}
}

@article{alkhatib2025prediction,
  title={Prediction via Shapley Value Regression},
  author={Alkhatib, Amr and Bresson, Roman and Bostr{\"o}m, Henrik and Vazirgiannis, Michalis},
  journal={arXiv preprint arXiv:2505.04775},
  year={2025}
}

\end{document}